\documentclass[11pt]{article}
\usepackage{acl}
\usepackage{times}
\usepackage{latexsym}
\usepackage[T1]{fontenc}
\usepackage[utf8]{inputenc}
\usepackage{microtype}
\usepackage{inconsolata}
\usepackage{booktabs}
\usepackage{amsmath}
\usepackage{tabularx}
\usepackage{dblfloatfix}
\usepackage{graphicx}
\usepackage{newunicodechar}
\newunicodechar{⁻}{\textsuperscript{-}}

\title{Recognized but Not Produced: A Generation Benchmark for Culturally Specific Kinship Terms}
\author{Sahil Pardasani and Madhusudan Singh \\ Blockchain Data Intelligence Lab\\
The Pennsylvania State University\\
University Park, PA, USA \\{mps6990@psu.edu}}

\begin{document}
\maketitle

\begin{abstract}
Current literature evaluates large language models (LLMs) on multilingual kinship understanding using multiple-choice benchmarks, treating it as a recognition problem. We instead prompt five open-weight LLMs to generate kinship terms in three non-Western languages (Hindi, Tamil, and Korean) across two communicative tasks, and pair this with a matched option-supported selection baseline. On identical relation--language cells, GPT-OSS-120B selects the correct term in 90.67\% of 75 valid cells but produces an accepted term in 36.00\% of the corresponding attempts; Llama-3.3-70B shows the same pattern (77.92\% versus 24.24\%). Since the four-option condition displays the candidate terms and does not require script production, the difference is interpreted as an evaluation-format gap rather than direct proof that lexical knowledge is intact. On explicitly specified L3 prompts, accuracy varies sharply, from GLM-5.1 at 72.29\% to Llama-3.3-70B at 24.24\%. The paternal-lineage advantage is language-specific: it is large in Hindi, but weak or reversed in Korean, while Tamil shared-term pairs provide a control for measurement variation. These results show that culturally specific kinship generation remains difficult even when the relationship is explicitly stated and motivate generation-based evaluation alongside multiple-choice testing.
\end{abstract}

\section{Introduction}
Significant efforts have expanded the multilingual capabilities of language models so that they can serve more people across languages. These efforts have encouraged major developers to train models on multilingual data. Although current models can respond fluently in many languages, they do not always generate culturally specific kinship terms correctly. Hindi, Korean, Tamil, Chinese, Arabic, Russian, Japanese, and Lithuanian are among the languages whose kinship systems make distinctions that English terms such as \textit{uncle} or \textit{grandmother} collapse. These distinctions can encode maternal or paternal lineage, relative age, marriage, and the speaker's position in the family. For example, in Hindi, a father's elder brother is called \textit{Tau/Tauji}, while a father's younger brother is called \textit{Chacha/Chachaji}. Models may translate both from English as a general term or default to the more frequent form even when the prompt specifies the relationship.

This pattern raises a broader question: even when a model can select a culturally specific term, can it reliably generate that term in a natural message? We study this question in Hindi, Korean, and Tamil using five open-weight models from different providers: GPT-OSS-120B \citep{agarwal2025gptoss}, Sarvam-M, Llama-3.3-70B, GLM-5.1 \citep{zeng2026glm}, and Kimi-K2.6. The models generate birthday greetings and wedding invitations in which the target relationship is explicitly specified.

Our contribution is a generation benchmark for culturally specific kinship terms together with a matched four-option baseline over the same relation--language cells. Selecting a displayed term and producing it in a complete message place different demands on a model. Because the multiple-choice condition supplies the candidate terms and removes spelling and native-script production, we call it \textit{option-supported selection} rather than treating it as direct proof of recognition. We first report generation and selection results and then examine whether paternal--maternal differences remain consistent across languages and models. This ordering connects the benchmark's overall performance differences with the narrower lineage analysis that follows.

\section{Related Work}

Kinship terminology is a well-established source of lexical gaps in multilingual resources. \citet{li2024lexgap} define a lexical gap as the absence of a single-word lexicalization for a concept, even when that concept can be expressed as a phrase. Using kinship as a case study, they show how English-centered resources can collapse gender-, age-, or lineage-specific concepts into broader categories. Their translation-based method improves lexical-gap detection and lexicalization generation over BabelNet and ChatGPT. In related work, \citet{khishigsuren2022kinship} construct a multilingual resource containing 198 concepts, 1,911 lexicalizations, and 37,370 lexical gaps across 699 languages. Their findings demonstrate the diversity of kinship systems and show that translation systems can introduce meaning-level errors when direct English equivalents do not exist.

Language structure and representation also affect multilingual generation. \citet{nzeyimana2024morph} study Kinyarwanda and argue that standard subword tokenization is insufficient when meaning is strongly encoded through morphology. Their morphology-aware translation framework models stems, affixes, parts of speech, and affix sets. For non-Roman scripts, \citet{husain2024romansetu} propose ROMANSETU, which uses romanization, continual pretraining, and instruction tuning to reduce tokenization fertility and improve multilingual performance. These studies suggest that errors involving culturally specific terms can reflect both lexical and representational difficulties.

Another line of work evaluates cultural alignment. \citet{li2024culturellm} introduce CultureLLM, using World Values Survey data and semantic augmentation to improve cultural awareness across nine cultures. \citet{alkhamissi2024cultural} compare model responses with survey data from Egypt and the United States and find that cultural alignment improves when models are prompted in the dominant language of the target culture. Together, these results show that multilingual fluency alone does not guarantee culturally appropriate behavior.

Our comparison also relates to work that separates generation from recognition or verification. \citet{davidson2026gvgap} show that language models often verify factual statements more reliably than they generate them and argue that the two capabilities should be evaluated independently. \citet{ranaldi2024empowering} study a related asymmetry during multilingual instruction tuning. Their CrossAlpaca models augment language-specific Alpaca data with bidirectional translation-following demonstrations from News Commentary and evaluate six languages on XQUAD, MLQA, and translated MMLU and BBH tasks. The added demonstrations improve over monolingual translated Alpacas, particularly on language-dependent question answering, but the gains remain below the English-trained baseline and vary with language, task, and translation direction. This work identifies training-time semantic alignment as one route to stronger cross-lingual behavior.

Culture-aware translation research further shows why literal translation is insufficient. \citet{yuan2026canmt} introduce CanMT, a benchmark covering 12 translation directions and evaluating contextual accuracy, cultural adaptation, functional equivalence, fidelity, and naturalness. \citet{conia2024xctranslate} focus on culturally nuanced entities and propose retrieval from multilingual knowledge graphs to improve cross-cultural translation.

Taken together, prior work establishes that kinship terminology varies substantially across languages, that multilingual representation can affect generation, and that models may fail to deploy available information during instruction following. However, it does not directly test whether recently released open-weight models can use culturally specific kinship terms in complete role-based messages when the relationship is explicitly given. We address this narrower gap by comparing repeated open generation with option-supported selection on matched relation--language cells and by examining how the resulting errors vary across languages and family relations.

\section{Methodology and Experimentation}

Our benchmark comprises 25 kinship relations that are lexically distinct in at least one of Hindi, Tamil, and Korean. It spans grandparents, parents' siblings and their spouses, siblings' children, and affinal relations. The source taxonomy contains 27 cross-lingual mappings. The experiments use 25 relationships; the generic Sister and Wife's sister entries are retained in the appendix only as reference mappings. Gold terms were initially compiled and checked by one native speaker per language. A second native speaker for each language independently reviewed the complete table, including the relation labels, term mappings, and listed variants. The second review confirmed all 81 language-specific entries (27 rows across three languages) without requesting changes, so no disagreement required adjudication.

\subsection{Prompt and Benchmark Construction}

We evaluate kinship generation in two communicative settings: birthday greetings and wedding invitations. These tasks were selected because they require the model not only to retrieve the appropriate kinship term, but also to use it naturally in socially meaningful discourse involving respect, affection, and politeness. The two templates are: ``Write a short message in \{lang\} wishing my \{rel\} a happy birthday'' and ``Write a short message in \{lang\} inviting my \{rel\} to my wedding.''

We construct prompts at two specificity levels. L1 replaces the relationship with an under specified English hypernym such as \textit{uncle}, \textit{aunt}, or \textit{grandmother}. This condition tests whether the model requests clarification, offers multiple culturally valid terms, or defaults to a high-frequency kinship term despite the unresolved ambiguity. L3 specifies the relationship completely, for example, \textit{father's elder brother} or \textit{husband's sister's husband}. This condition tests whether the model can generate the correct culturally specific term once the underlying relationship is made explicit. Elder brother is the only relationship without an L1 formulation. It contributes four L3-only relation--language--perspective cells because Korean has male- and female-speaker variants.

Relations whose correct term depends on the speaker's position include a perspective prefix, such as ``I am the wife.'' Korean relations whose lexical realization depends on speaker gender are divided into male and female-speaker variants. Each benchmark record stores a unique prompt ID, relationship label, specificity level, task type, target language, speaker perspective, expected kinship term, and final prompt text.

Crossing 25 relations with three languages, two tasks, and two specificity levels, while accounting for L3-only relations and Korean speaker-gender variants, yields 300 unique prompts per model: 146 at L1 and 154 at L3. Each prompt is run three times to reduce the influence of stochastic decoding and to assess consistency across generations. This produces 900 responses per model and 4,500 responses in total. The design contains 73 L1 cells and 77 L3 cells. The eight L3-only task prompts correspond to 24 responses across three runs. We do not convert the six L3 attempts in a relation--language--perspective cell into a majority-vote label: cell accuracy is the mean of all six binary outcomes, so a three-to-three split remains 0.50 and requires no tie-breaking rule.

\subsection{Evaluation Protocol}

We evaluate outputs using a frozen rule-based matcher containing the accepted native-script kinship terms together with enumerated spelling, spacing, and honorific variants. A response is counted as correct when one of these strings appears as a literal substring in the generated text. All 71 expected-term--language keys observed in the five-model analysis are covered by the strict native-script map, so the scorer's generic fallback is not used for the reported 4,500 responses. The matcher does not perform Unicode normalization, morphological or semantic analysis, or word-boundary enforcement, and it does not require the complete message to reproduce a reference sentence.

The score is therefore a test of accepted-term production rather than a full evaluation of grammar or message quality. A failed match can include a wrong relation, Romanization, the wrong script or language, a message that omits the kinship term, or a valid regional or inflected form not covered by the inventory. We use a script-level breakdown in the failure analysis and discuss the need for broader human validation in the Limitations section.

Tamil does not lineage-mark several relations: father's father and mother's father are both \textit{thaatha}, both grandmothers are \textit{paatti}, and the same listed terms are used for siblings' children across lineage. Four of the five mirrored paternal--maternal pairs therefore share the same gold token in Tamil. These cells serve as a control for measurement variation in the lineage analysis.

\subsection{Option-Supported Selection Baseline and Decoding Setup}

We generate 77 four-option multiple-choice items, one for each fully specified relation--language--perspective cell. Option-supported selection is evaluated for GPT-OSS-120B and Llama-3.3-70B. The remaining three models are evaluated only on generation.

During evaluation, the English stem and instruction were followed by four options in the target language's native script, exactly as stored in the recognition data. The model returned only a Latin option letter: A, B, C, or D. All four surface forms are unique within an item, and the gold answer appears exactly once. The distractors are broad inventory terms rather than deliberately close relation or lineage alternatives. A seeded shuffle places the answer at A in 13 items, B in 24, C in 21, and D in 19; the positions are therefore randomized but not exactly balanced. Generation uses temperature 0.7, top-$p$ 0.95, a 1,024-token maximum, and three runs per prompt. Selection uses temperature 0 and one run per item.

The parser accepts the first standalone letter A, B, C, or D. A blank or unparseable response receives no selection label and is excluded from the matched-cell comparison; treating such responses as incorrect is also reported as a sensitivity result. GPT-OSS-120B has 75 valid selection outputs after two blank responses are excluded, while Llama-3.3-70B has all 77. GPT therefore scores 68/75 (90.67\%) on valid outputs and 68/77 (88.31\%) when the two blanks are counted as incorrect. Four Korean speaker-perspective labels are normalised before matching the GPT selection and generation cells.

Appendix~\ref{sec:recognition-examples} reproduces two released items and their responses. Evaluation used English stems and instructions with native-script options. Only the appendix display is transliterated; the released files preserve the original scripts.

\begin{table}[t]
\centering
\small
\resizebox{\columnwidth}{!}{%
\begin{tabular}{lrrr}
\toprule
Model & Correct & Acc. (\%) & Cell-bootstrap 95\% CI \\
\midrule
GLM-5.1       & 334/462 & 72.29 & [64.94, 79.22] \\
Kimi-K2.6     & 225/462 & 48.70 & [40.04, 57.36] \\
Sarvam-M      & 187/462 & 40.48 & [32.03, 49.13] \\
GPT-OSS-120B  & 162/462 & 35.06 & [26.62, 43.72] \\
Llama-3.3-70B & 112/462 & 24.24 & [16.45, 32.47] \\
\bottomrule
\end{tabular}
}
\caption{L3 generation accuracy over 77 cells per model. Intervals resample whole relation--language--perspective cells, keeping all six attempts together.}
\label{tab:overall}
\end{table}

\section{Findings}
\subsection{Overall generation accuracy}
Table~\ref{tab:overall} shows a wide spread and a low ceiling on explicitly specified L3 prompts. GLM-5.1 reaches 72.29\%, followed by Kimi-K2.6 at 48.70\%, Sarvam-M at 40.48\%, GPT-OSS-120B at 35.06\%, and Llama-3.3-70B at 24.24\%. The intervals are computed over relation--language--perspective cells so that repeated generations from the same prompt remain together.

Task-level differences are small relative to the variation between models. Across all five models, accepted-term accuracy is 44.07\% for birthday prompts and 44.24\% for wedding invitations. Table~\ref{tab:tasks} shows that the direction of the difference is not consistent across models. Among L3 failures, 135 birthday responses and 119 invitation responses contain no character from the requested script. Full breakdowns by model, language, task, and relation are included in the release.

\begin{table}[t]
\centering
\small
\begin{tabular}{lrr}
\toprule
Model & Birthday & Wedding invitation \\
\midrule
GLM-5.1       & 73.59 & 71.00 \\
Kimi-K2.6     & 47.19 & 50.22 \\
Sarvam-M      & 41.13 & 39.83 \\
GPT-OSS-120B  & 35.06 & 35.06 \\
Llama-3.3-70B & 23.38 & 25.11 \\
\bottomrule
\end{tabular}
\caption{L3 accepted-term accuracy (\%) by communicative task. Each model contributes 231 responses per task.}
\label{tab:tasks}
\end{table}

\begin{table}[t]
\centering
\small
\resizebox{\columnwidth}{!}{%
\begin{tabular}{lrrrr}
\toprule
Model & Cells & Selection & Generation & Gap [95\% CI] \\
\midrule
GPT-OSS-120B  & 75 & 90.67 & 36.00 & 54.67 [45.11, 64.00] \\
Llama-3.3-70B & 77 & 77.92 & 24.24 & 53.68 [40.91, 65.58] \\
\bottomrule
\end{tabular}
}
\caption{Option-supported selection versus mean L3 generation accuracy (\%) on matched cells. Generation averages the six attempts in each cell.}
\label{tab:recognition}
\end{table}

Option-supported selection is substantially higher than open generation for both evaluated models. GPT-OSS-120B selects the correct option in 68 of 75 valid cells, while Llama-3.3-70B is correct in 60 of 77. On the same cells, generation reaches 36.00\% and 24.24\%, respectively. The difference is large, but the selection condition displays the candidate terms and avoids spelling, native-script production, and full-message generation. The comparison therefore shows an evaluation-format gap rather than proving that lexical knowledge is fully intact.

Both selection results are above the 25\% four-choice chance baseline (one-sided exact binomial tests, $p<.001$). Because the shuffled answer positions are not exactly balanced, we also average accuracy across positions A, B, C, and D. The position-macro scores are 91.07\% for GPT-OSS-120B and 78.29\% for Llama-3.3-70B, close to their ordinary valid-output accuracies of 90.67\% and 77.92\%. The release provides the counts for every model and answer position.

The gap also varies by language. For GPT-OSS-120B it is 63.19 points in Hindi, 58.02 in Korean, and 42.36 in Tamil. For Llama-3.3-70B it is 36.67 in Hindi, 100.00 in Korean, and 20.67 in Tamil. Llama's Korean result is strongly affected by its script-production failure discussed in Section~\ref{sec:failures}.

\begin{table}[t]
\centering
\small
\begin{tabular}{lrrr}
\toprule
Pair & Hindi & Korean & Tamil \\
\midrule
Grandfather       & +43.3 & +36.7 & +13.3$^{\dagger}$ \\
Grandmother       & +50.0 & +40.0 & +23.3$^{\dagger}$ \\
Parent's sister   & +26.7 & $-43.3$ & +16.7 \\
Sibling's daughter& +46.7 & $-13.3$ & +3.3$^{\dagger}$ \\
Sibling's son     & +50.0 & 0.0 & $-3.3^{\dagger}$ \\
\midrule
Signed mean       & +43.3 & +4.0 & +9.2$^{\dagger}$ \\
\bottomrule
\end{tabular}
\caption{Paternal minus maternal L3 accuracy in percentage points on five mirrored pairs, pooled over five models ($n=30$ per cell). $\dagger$ marks Tamil control pairs sharing a gold token across both sides.}
\label{tab:lineage}
\end{table}

GPT-OSS-120B returned \textit{dadiji} for a maternal grandmother, \textit{halmeoni} for \textit{oe-halmeoni}, and \textit{bhatiji} for \textit{bhanji}, yet was correct on every run for father's mother. Table~\ref{tab:lineage} shows that the paternal--maternal difference is large and consistent in Hindi (mean $+43.3$ points) but only $+4.0$ in Korean, where it reverses for parent's sister and sibling's daughter. The Tamil controls have a signed mean of $+9.2$ points and a mean absolute gap of 10.83 points even though four pairs share the same gold token. This provides a practical reference for sampling and scorer variation.

The pooled means do not describe every model equally. Table~\ref{tab:lineage-model} therefore reports the same signed mean separately by model. Hindi ranges from $+3.3$ points for GLM-5.1 to $+90.0$ for GPT-OSS-120B; Korean ranges from $-6.7$ to $+16.7$; and Tamil ranges from $-16.7$ to $+30.0$. We consequently treat the lineage result as a model- and language-specific pattern rather than one consistent multilingual bias.

\begin{table}[t]
\centering
\small
\begin{tabular}{lrrr}
\toprule
Model & Hindi & Korean & Tamil \\
\midrule
GLM-5.1       & +3.3 & 0.0 & +6.7 \\
Kimi-K2.6     & +13.3 & +10.0 & +10.0 \\
Sarvam-M      & +43.3 & +16.7 & +23.3 \\
GPT-OSS-120B  & +90.0 & $-6.7$ & +30.0 \\
Llama-3.3-70B & +66.7 & 0.0 & $-16.7$ \\
\bottomrule
\end{tabular}
\caption{Signed paternal-minus-maternal L3 mean by model and language over the five mirrored pairs. Tasks and repeated runs remain grouped within each relation cell.}
\label{tab:lineage-model}
\end{table}

\subsection{Failure modes}
\label{sec:failures}

Llama-3.3-70B emitted no Hangul in 162 of 162 Korean L3 responses while producing Devanagari and Tamil normally. These responses correspond to all 27 Korean L3 cells. All 162 are non-empty and contain ASCII characters only; 120 explicitly include an English ``Translation'' field and 76 include a ``Note'' field. The outputs typically mix Romanized Korean with English explanations rather than producing Korean script. This systematic script failure contributes strongly to the model's Korean generation score.

Across all 4,500 generations, 1,742 responses pass and 2,758 fail the accepted-term matcher. We apply a second deterministic surface screen to describe what these failures contain. Of the 2,758 failures, 1,597 (57.90\%) contain at least one different target-language token from the accepted inventory, 679 (24.62\%) contain target-script characters but no token from the inventory, and 469 (17.01\%) contain no character from the requested script. The last group includes 34 responses containing the expected Romanized form. The remaining 13 failures (0.47\%) contain target-script text and the expected Romanization but not an accepted native-script form.

The L3-only pattern is similar: 672 of 1,290 failures (52.09\%) contain another inventory token, 351 (27.21\%) contain target-script text but no inventory token, 254 (19.69\%) lack the target script, and 13 (1.01\%) contain the expected Romanization alongside target-script text. Among the 1,020 L3 passes, 312 (30.59\%) also contain at least one additional inventory token. These are literal surface categories rather than semantic judgments: another token can occur in an explanation or natural family context, and the no-inventory category can include omitted terms, valid unlisted variants, or scorer false negatives. Row-level categories and summaries by model, language, task, and relation are included in the release.

Affinal relations are among the most difficult categories, but the pattern is not uniform. L3 accuracy is 4.44\% for husband's sister's husband, 6.67\% for younger sister's husband, 15.56\% for elder sister's husband, 18.89\% for husband's sister, and 20.00\% for wife's brother. Elder brother's wife is notably higher at 58.33\%.

The script-level failure pattern also differs by model and language. Among L3 failures, the number lacking the requested script is 0/41, 0/37, and 0/50 for GLM-5.1 in Hindi, Korean, and Tamil; 47/68, 15/83, and 22/86 for Kimi-K2.6; 1/82, 7/113, and 0/80 for Sarvam-M; 0/103, 0/94, and 0/103 for GPT-OSS-120B; and 0/91, 162/162, and 0/97 for Llama-3.3-70B. This breakdown separates script-production failures from target-script outputs that fail the accepted-term inventory, although a manual semantic error analysis is still required.

\section{Conclusion}
We introduced a generation benchmark for culturally specific kinship terms covering 25 tested relations, three languages, two communicative tasks, and five open-weight models over 4,500 generations. The results show a wide spread in explicitly specified L3 generation, from 72.29\% for GLM-5.1 to 24.24\% for Llama-3.3-70B. A matched option-supported selection baseline is substantially higher for GPT-OSS-120B and Llama-3.3-70B, showing that performance depends strongly on whether a model selects a displayed term or produces it in a complete message.

The deficit is language- and relation-specific. The paternal--maternal difference is large in aggregate for Hindi but varies substantially by model, while Korean is weak or reversed and Tamil shared-term pairs expose measurement variation even when the expected token is identical. Llama-3.3-70B also produces no Hangul in any of its 162 Korean L3 responses. Together, these findings show why multilingual kinship should be evaluated through generation rather than multiple-choice accuracy alone.

\section*{Limitations}
Three languages from two families and 25 tested relations in a single standard register cannot support claims about kinship generation in general; regional dialects and evolving contemporary usage may license terms our lists reject.

Scoring is token-presence based. It can miss legitimate inflections and can credit a token inside an explanation even when the address term is wrong. The surface audit cannot distinguish a wrong relation or language, omission, valid unlisted variant, clarification, or malformed output, nor determine whether a passing term is used appropriately. Without a stratified human audit, we cannot report the matcher's false-positive or false-negative rate.

Two native speakers per language reviewed the gold inventory, and the second reviewers confirmed all 81 language-specific entries without changes. This was table-level review rather than blinded item-level annotation, so we report 100\% confirmation but no agreement coefficient. Slash-marked alternatives can also represent different linguistic conditions and were not tested as separate randomized items.

L1 prompts admit several valid answers but are scored against one gold term, so they can under-report performance. We therefore use fully specified L3 as the primary condition and exclude L1 from the headline and selection--generation tables.

Option-supported selection covers two of five models, uses broad inventory distractors, always contains the answer, and permits elimination without spelling, script production, or message composition. It also uses one run at temperature 0 rather than three at 0.7. The comparison therefore does not isolate retrieval and is a task-format difference, not a direct measure of recognition.

The panel differs in size, architecture, provider, and language support and has no smaller-model or language-specialist baseline. Equivalent general proficiency in the three languages was not established, so cross-model differences are descriptive rather than controlled evidence about scale or cultural knowledge.

The original LaTeX setup did not reliably render Devanagari and Hangul, so the appendix uses transliteration.

Finally, we do not observe the models' training corpora and therefore make no causal claim about the reported patterns.

\section*{Future Work}
Future evaluation should add close relation- and lineage-level distractors, balanced answer positions, answer-absent items, free recall, script controls, and a condition that supplies the correct term for use in a complete message. Selection should cover all evaluated models, including smaller and language-specialist baselines, with target-language proficiency assessed independently.

L1 should distinguish clarification, multiple valid alternatives, and unjustified defaulting. Composite gold entries should become condition-specific items for speaker gender, relative age, dialect, and related distinctions. Broader validation should document regional varieties, reviewer backgrounds, adjudication, and term provenance. A blinded, stratified audit of passes and failures should report semantic errors by model, language, relation, and task, calculate agreement, and estimate scorer error rates.

We also plan a controlled paired study of retrieval support. In a small exploratory pilot, a graph mapping native kinship terms to English relations was exposed to a model as a tool over the Model Context Protocol. On a locally served 20B model outside our evaluation set, the tool-augmented setting produced \textit{tauji} (father's elder brother) in Hindi and \textit{oe-halmeoni} (mother's mother) in Korean where the base model did not. These two examples are illustrative only, are not included in the benchmark or its statistics, and do not show that the tool solves the task; a matched multi-relation comparison is left to future work.

\section*{Data and Code Availability}
{\raggedright
The data, code, and documented offline reproduction workflow are available at \url{https://github.com/sahilpardasani/kinship-generation-benchmark}.
Human output-audit records are not included because a stratified annotation of model outputs has not yet been conducted.
\par}

\section*{Ethics Statement}
This work evaluates relationship terms in Hindi, Tamil, and Korean. Such terms are intertwined with social structure, gender roles, and communal identity; incorrect generation can contribute to cultural misalignment and linguistic homogenization. Gold terms and accepted variants were independently reviewed by two native speakers per language, and the second reviewers confirmed all 81 language-specific entries without requesting changes. No private or personally identifiable information was collected, and all model inputs were synthetic role-based prompts.

\section*{Acknowledgments}
Artificial intelligence was used as part of the experiments and to assist with LaTeX formatting of the camera-ready manuscript and the identification of grammatical, typographical, and presentation errors. The authors reviewed and verified the experimental outputs, analyses, citations, and manuscript changes and take full responsibility for the methods, results, and content of the paper.

\bibliography{custom}

\appendix 
\section{Count Reconciliation}
\begin{center}
\centering
\small
\begin{tabular}{lr}
\toprule
Quantity & Value \\
\midrule
Tested relationships & 25 \\
Reference-only mappings & Sister; Wife's sister \\
Unique prompts per model & 300 \\
L1 / L3 task prompts & 146 / 154 \\
L1 / L3 cells & 73 / 77 \\
Runs per prompt & 3 \\
Responses per model & 900 \\
L3-only relationship / cells & Elder brother / 4 \\
L3-only prompts / responses & 8 / 24 \\
Models analyzed & 5 \\
Total generations & 4,500 \\
Selection items per model & 77 \\
GPT valid selection items & 75 \\
Llama valid selection items & 77 \\
\bottomrule
\end{tabular}
\captionof{table}{Reconciliation of the main denominators used in the paper.}
\label{tab:counts}
\end{center}

\subsection{Model and run documentation}
Table~\ref{tab:model-documentation} reports the exact identifiers recorded by the experiment configuration. All five models use the generation templates in Section~3.1 and contribute 900 generation responses. Only GPT-OSS-120B and Llama-3.3-70B are included in option-supported selection. Per-call timestamps, provider labels, error fields, and analysis membership are available in \texttt{model\_run\_metadata.csv} and the row-level audit files.

\begin{center}
\scriptsize
\setlength{\tabcolsep}{2.5pt}
\begin{tabularx}{\columnwidth}{@{}lXrr@{}}
\toprule
Model & Recorded identifier & Gen. & Valid sel. \\
\midrule
GLM-5.1 & z-ai/glm-5.1 & 900 & -- \\
Kimi-K2.6 & moonshotai/kimi-k2.6 & 900 & -- \\
Sarvam-M & sarvamai/sarvam-m & 900 & -- \\
GPT-OSS-120B & openai/gpt-oss-120b & 900 & 75 \\
Llama-3.3-70B & meta/llama-3.3-70b-instruct & 900 & 77 \\
\bottomrule
\end{tabularx}
\captionof{table}{Exact configured model identifiers and analyzed response counts.}
\label{tab:model-documentation}
\end{center}

\subsection{Recognition examples}
\label{sec:recognition-examples}
These examples come from the released data. Evaluation used English questions and instructions with native-script options; models returned Latin option letters. Options below are transliterated only for this LaTeX rendering.

\medskip
\noindent\textbf{Example 1: Hindi}\hfill\textit{Gold: C}

\smallskip
{\setlength{\fboxsep}{4pt}%
\noindent\fbox{\begin{minipage}{0.92\columnwidth}
\small
Which kinship term in Hindi correctly refers to my father's younger brother? Answer with only the option letter.

\medskip
A. \textit{bhatija}\\
B. \textit{tau}\\
C. \textit{chacha}\\
D. \textit{mausi}

\medskip
\textbf{Answer with only one option letter: A, B, C, or D.}
\end{minipage}}}

\smallskip
\noindent\textit{Recorded responses:} GPT-OSS-120B: ``C'' (correct); Llama-3.3-70B: ``C.'' (parsed as C; correct).\\
\noindent\textit{Dataset ID:} father\_younger\_brother\_\_REC\_\_HI.

\smallskip
\noindent\textbf{Example 2: Tamil}\hfill\textit{Gold: C}

\smallskip
{\setlength{\fboxsep}{4pt}%
\noindent\fbox{\begin{minipage}{0.92\columnwidth}
\small
Which kinship term in Tamil correctly refers to my father's younger brother? Answer with only the option letter.

\medskip
A. \textit{periyamma}\\
B. \textit{naathanar}\\
C. \textit{chithappa}\\
D. \textit{anni}

\medskip
\textbf{Answer with only one option letter: A, B, C, or D.}
\end{minipage}}}

\smallskip
\noindent\textit{Recorded responses:} GPT-OSS-120B: ``B'' (incorrect); Llama-3.3-70B: ``C.'' (parsed as C; correct).\\
\noindent\textit{Dataset ID:} father\_younger\_brother\_\_REC\_\_TA.

\clearpage
\onecolumn
\section{Kinship Taxonomy}
Table~\ref{tab:kinship-taxonomy} preserves the original 27-row taxonomy structure while reporting the actual Romanized expected-term keys used for the 25 tested relationships. Sister and Wife's sister are reference-only mappings and are not included in the prompt or accuracy counts. Slash notation is not a single semantic category: it marks accepted address alternatives in rows such as Hindi elder brother, speaker-gender-conditioned forms in Korean elder brother, and relative-age-conditioned forms in some Tamil sibling rows. These alternatives were not evaluated as separate randomized conditions. \begin{center}
\centering
\small
\setlength{\tabcolsep}{3.5pt}
\renewcommand{\arraystretch}{0.96}
\begin{tabularx}{\textwidth}{@{}>{\raggedright\arraybackslash}X >{\raggedright\arraybackslash}X >{\raggedright\arraybackslash}X >{\raggedright\arraybackslash}X@{}}
\toprule
\textbf{English relationship} & \textbf{Hindi term} & \textbf{Tamil term} & \textbf{Korean term} \\
\midrule
Father's father & dadaji & thatha & harabeoji \\
Father's mother & dadi (dadiji) & paati & halmeoni \\
Father's elder brother & tau (tauji) & periyappa & keun-abeoji \\
Father's elder brother's wife & tai (taiji) & periyamma & keun-eomeoni \\
Father's younger brother & chacha (chachaji) & chithappa & samchon \\
Father's younger brother's wife & chachi (chachiji) & chithi & jageun-eomeoni \\
Father's sister & bua (buaji) & athai & gomo \\
Father's sister's husband & phupha & mama / attan & gomobu \\
Brother's daughter & bhatiji & marumagal & jokattal \\
Brother's son & bhatija & marumagan & joka \\
Mother's father & nanaji & thatha & oe-harabeoji \\
Mother's mother & nani (naniji) & paati / ammayee & oe-halmeoni \\
Mother's brother & mama (mamaji) & mama / maaman & oe-samchon \\
Mother's brother's wife & mami (mamiji) & mami & oe-sungmo \\
Mother's sister & mausi (mausiji) & chithi / periyamma & imo \\
Mother's sister's husband & mausa & mama / chithappa & imobu \\
Sister's daughter & bhanji & marumagal & jokattal / saengjilnyeo \\
Sister's son & bhanja & marumagan & joka / saengjil \\
Elder brother & bhaiya / bhai & anna & hyung (m) / oppa (f) \\
Elder brother's wife & bhabhi & anni & hyeongsu (m) / olke (f) \\
Sister & Didi / Badi Bahen & Akka & Nuna (m) / Eonni (f) \\
Sister's husband & jija (jijaji) & machan / attan & maehyeong \\
Younger sister's husband & bahanoi & machan & maeje \\
Husband's sister & nanad & nathanar & sinui \\
Husband's sister's husband & nandoi & machan / attan & seobang-nim \\
Wife's brother & sala & machan / maamaan & cheonam \\
Wife's sister & Saali & Kozhundhiyal & Cheohyeong / Cheoje \\
\bottomrule
\end{tabularx}
\captionof{table}{Romanized expected-term keys for the 25 tested relationships, with two reference-only mappings retained from the original taxonomy. The complete native-script variants used by the matcher are provided in the release; regional and orthographic variation is discussed in the Limitations section.}
\label{tab:kinship-taxonomy}
\end{center}

\end{document}